\documentclass[sigconf]{acmart}

\usepackage{xspace}
\usepackage{graphicx}
\usepackage{bm}
\usepackage{marvosym}
\usepackage[ruled,linesnumbered]{algorithm2e}

\newcommand{\paratitle}[1]{\vspace{1.5ex}\noindent\textbf{#1}}
\newcommand{\ie}{\emph{i.e.,}\xspace}

\newcommand{\eg}{\emph{e.g.,}\xspace}

\newcommand{\ignore}[1]{}

\newcommand{\framework}{CFCRS\xspace}

\copyrightyear{2023} 
\acmYear{2023} 
\setcopyright{acmlicensed}\acmConference[KDD '23]{Proceedings of the 29th ACM SIGKDD Conference on Knowledge Discovery and Data Mining}{August 6--10, 2023}{Long Beach, CA, USA}
\acmBooktitle{Proceedings of the 29th ACM SIGKDD Conference on Knowledge Discovery and Data Mining (KDD '23), August 6--10, 2023, Long Beach, CA, USA}
\acmPrice{15.00}
\acmDOI{10.1145/3580305.3599387}
\acmISBN{979-8-4007-0103-0/23/08}

\begin{document}

\title{Improving Conversational Recommendation Systems via Counterfactual Data Simulation}


\author{Xiaolei Wang$^{\dagger}$}
\affiliation{%
    \institution{Gaoling School of Artificial Intelligence, Renmin University of China}
    \city{Beijing}
    \country{China}
}
\email{wxl1999@foxmail.com}

\author{Kun Zhou$^{\dagger}$}
\affiliation{%
    \institution{School of Information, Renmin University of China}
    \city{Beijing}
    \country{China}
}
\email{francis_kun_zhou@163.com}

\author{Xinyu Tang$^{\dagger}$}
\affiliation{%
    \institution{Gaoling School of Artificial Intelligence, Renmin University of China}
    \city{Beijing}
    \country{China}
}
\email{txy20010310@163.com}

\author{Wayne Xin Zhao$^{\dagger}$\textsuperscript{\Letter}}
\affiliation{
    \institution{Gaoling School of Artificial Intelligence, Renmin University of China}
    \city{Beijing}
    \country{China}
}
\email{batmanfly@gmail.com}
\thanks{$\dagger$ Beijing Key Laboratory of Big Data Management and Analysis Methods.}
\thanks{\Letter\ Corresponding author.}

\author{Fan Pan, Zhao Cao}
\affiliation{%
    \institution{Huawei Poisson Lab}
    \city{Beijing}
    \country{China}
}
\email{panfan3@huawei.com}
\email{caozhao1@huawei.com}

\author{Ji-Rong Wen$^{\dagger}$}
\affiliation{%
    \institution{Gaoling School of Artificial Intelligence, Renmin University of China}
    \city{Beijing}
    \country{China}
}
\email{jrwen@ruc.edu.cn}

\renewcommand{\shortauthors}{Wang, et al.}

\begin{abstract}
Conversational recommender systems~(CRSs) aim to provide recommendation services via natural language conversations.  
Although a number of approaches have been proposed for developing capable CRSs, they typically rely on sufficient training data for training. 
Since it is difficult to annotate recommendation-oriented dialogue datasets, existing CRS approaches often suffer from the issue of \emph{insufficient training} due to the scarcity of training data.

To address this issue, in this paper, we propose a \textbf{C}ounter\textbf{F}actual data simulation approach for \textbf{CRS}, named \textbf{CFCRS}, to alleviate the issue of data scarcity in CRSs.
Our approach is developed based on the framework of \emph{counterfactual data augmentation}, which gradually incorporates the rewriting to the user preference from a real dialogue without interfering with the entire conversation flow. 
To develop our approach, we characterize user preference and organize the conversation flow by the entities involved in the dialogue, and design a multi-stage recommendation dialogue simulator based on a conversation flow language model. 
Under the guidance of the learned user preference and dialogue schema, the flow language model can produce \emph{reasonable, coherent} conversation flows, which can be further realized into complete dialogues.
Based on the simulator, we perform the intervention at the representations of the interacted entities of target users, and design an adversarial training method with a curriculum schedule that can gradually optimize the data augmentation strategy. 
Extensive experiments show that our approach can consistently boost the performance of several competitive CRSs, and outperform other data augmentation methods, especially when the training data is limited.
Our code is publicly available at \textcolor{blue}{\url{https://github.com/RUCAIBox/CFCRS}}.
\end{abstract}

\ccsdesc[500]{Information systems~Recommender systems}

\keywords{Conversational Recommender System; Counterfactual Data Augmentation; Prompt Learning}

\maketitle

\section{Introduction}

The recent success of conversational intelligence~\cite{gao2018neural,chen2017survey} has empowered a more convenient way for information seeking by \emph{conversational recommender systems}~(CRSs)~\cite{jannach2021survey,gao2021advances,christakopoulou2016towards}, which aims to provide high-quality recommendation service through multi-turn natural language conversations.
Typically, a CRS recommends the suitable items that satisfy the user need via \emph{a recommender module}, and generates the proper response based on the conversation context and predicted items via \emph{a conversation module}. 
These two modules are systematically integrated to fulfill the information-seeking task. 

To develop capable CRSs, various approaches have been proposed in the literature~\cite{zhou2022cr,ren2022variational,li2022user} based on deep neural networks. 
In particular, the powerful Transformer network~\cite{chen2019towards} and pre-trained language models~(PLM)~\cite{wang2022recindial} have largely raised the performance bar on conversational recommendation.
These approaches rely on high-quality recommendation-oriented conversation data for model training. 
While it is difficult to manually create large-scale CRS datasets, which require well-trained annotators to generate \emph{coherent, diverse} conversation flow in an information-seeking scenario~\cite{liu2020towards,zhou2020towards}. 
Therefore, existing CRS datasets~\cite{li2018towards,liu2020towards} are often limited in data size, lacking sufficient coverage of diverse information needs and user preferences.   
To alleviate this issue, existing studies incorporate external data (\eg knowledge graphs~\cite{chen2019towards,zhou2020improving}) and model resources (\eg DialoGPT~\cite{zhang2020dialogpt}) to reduce the demand for training data in developing a capable CRS. 

Despite the performance improvement, the fundamental issue of \emph{insufficient training} in existing CRSs has not been well addressed due to the scarcity of training datasets. 
As a general solution to data scarcity, data augmentation techniques~\cite{shorten2019survey,feng2021survey,wen2020time} have been widely applied in a variety of tasks, which either use heuristic strategies~\cite{DBLP:conf/naacl/WangB18,DBLP:conf/emnlp/WeiZ19} or learnable models~\cite{DBLP:conf/nips/XieDHL020,DBLP:conf/acl/SinghGR18} for enlarging the data size. 
However, it is challenging to augment high-quality recommendation-oriented dialogues (short as \emph{recommendation dialogues}) with automatic approaches, since it needs to mimic the interactive information-seeking process via a reasonable, coherent conversation flow.  
To be \emph{reasonable}, the conversation scenario should be designed with meaningful user needs and suitable item recommendations, which conform to the factual information in the given domain. 
To be \emph{coherent}, the augmented user preference should be consistent throughout the whole conversation, which should be well clarified and maintained as the conversation progresses. 
Considering these difficulties, existing work~\cite{hou2018sequence,kumar2019submodular} that uses specific rewriting strategies cannot generate high-quality CRS datasets.

To enhance the reasonableness and coherence of the conversation flow, we take a holistic perspective to develop the augmentation approach for recommendation dialogues by gradually incorporating the rewriting or adaptation into a real dialogue. 
Specially, each rewriting is expected to be carefully controlled without interfering with the entire conversation flow. 
Indeed, such intuition can be well fit into the framework of \emph{counterfactual data augmentation}~\cite{goyal2019counterfactual,pitis2020counterfactual,liu2021counterfactual}, which incorporates counterfactual learning for augmenting the limited data. 
In this setting, the essence of our approach is to answer the key question: ``\emph{What the dialogues would be if we intervene on the observed user preference?''}, where user preference is considered to be the most important factor to determine a conversation flow.    
To instantiate it, we consider characterizing user preference and organizing the conversation flow by the entities involved in the dialogue (\eg movie actors and genres). 
Further, our rewriting strategy is implemented by a learnable edit function, which can produce informative edits to the entity representations for improving the recommendation module. 
In this way, the original user preference is gradually revised and finally reaches the level that a high-quality yet different conversation is augmented.  

To this end, in this paper, we present the proposed \textbf{C}ounter\textbf{F}actual data simulation approach for \textbf{CRS}, named \textbf{CFCRS}, for alleviating the issue of data scarcity in CRSs.  
Our core idea is to leverage counterfactual learning to augment user preference and then employ the augmented user preference to simulate conversation data. 
Specifically, we design a recommendation dialogue simulator that can generate \emph{reasonable, coherent} conversations for two target users. To guarantee the quality of the simulated conversations, we design a flow language model to generate the conversation flow, which is guided by the learned user preference and dialogue sketch. 
Based on the dialogue simulator, we perform the intervention at the representations of the interacted entities of target users, and design an adversarial training method with a curriculum schedule that can gradually optimize the edit function towards an improved recommendation capacity of CRSs.

To the best of our knowledge, it is the first time that \textit{counterfactual data simulation} has been utilized to improve CRS models.
Our proposed framework is agnostic to model implementations, hence is general to various CRS methods.
To evaluate the effectiveness of our approach, we evaluate its performance with several representative CRS models on two public CRS datasets.
Experimental results show that our approach can consistently boost the performance of these models, and outperform other data augmentation methods, especially when the training data is limited.
\section{Related Work}

In this section, we summarize the related work as follows.

\paratitle{Conversational Recommender System.}
Conversational recommender systems~(CRSs) aim to provide recommendation services through conversational interactions.
One line of work~\cite{lei2020estimation, shang2023multi, zhou2020leveraging} relies on pre-defined interactive actions (\eg asking preferences about item attributes or making recommendations) and hand-crafted templates to converse with users.
They mainly focus on capturing user preferences and giving accurate recommendations within as few turns as possible.
Another line of work~\cite{li2018towards, chen2019towards, zhou2020improving} focuses on interacting with users through more free-form natural language conversations.
They aim to capture the preferences from the conversation context and then generate the recommended items with persuasive responses.
The above CRS methods are mostly developed by deep neural networks, which require sufficient high-quality data for training.
However, it is expensive to annotate high-quality CRS examples and existing datasets are generally limited in scale.
To address it, external resources like knowledge graphs~\cite{zhou2020improving,chen2019towards} and reviews~\cite{lu2021revcore,zhou2022c2} have been introduced to enrich the datasets.
However, the fundamental problem of insufficient training examples has not been well solved.
In this work, we aim to solve the data scarcity problem via counterfactual data simulation.


\paratitle{Counterfactual Data Augmentation.}
Counterfactual data augmentation~\cite{zmigrod2019counterfactual,pitis2020counterfactual,liu2021counterfactual} focuses on generating unrecorded counterfactual examples from the real ones.
Recently, it has been leveraged to alleviate the data scarcity problem and can improve the performance and robustness of deep neural networks.
For example, in recommender systems, CASR~\cite{wang2021counterfactual} proposes to generate counterfactual user behavior sequences based on the real ones to supply the data for training sequential recommendation models.
While for open-domain dialogue generation, CAPT~\cite{ou2022counterfactual} uses counterfactual inference to automatically augment high-quality responses with different semantics to solve the one-to-many problem.
In this work, we apply counterfactual data augmentation to the CRS task, aiming to obtain sufficient high-quality data.
We also propose a curriculum learning strategy to gradually optimize the data augmentation strategy for CRSs.

\begin{figure*}
  \centering
  \includegraphics[width=\textwidth]{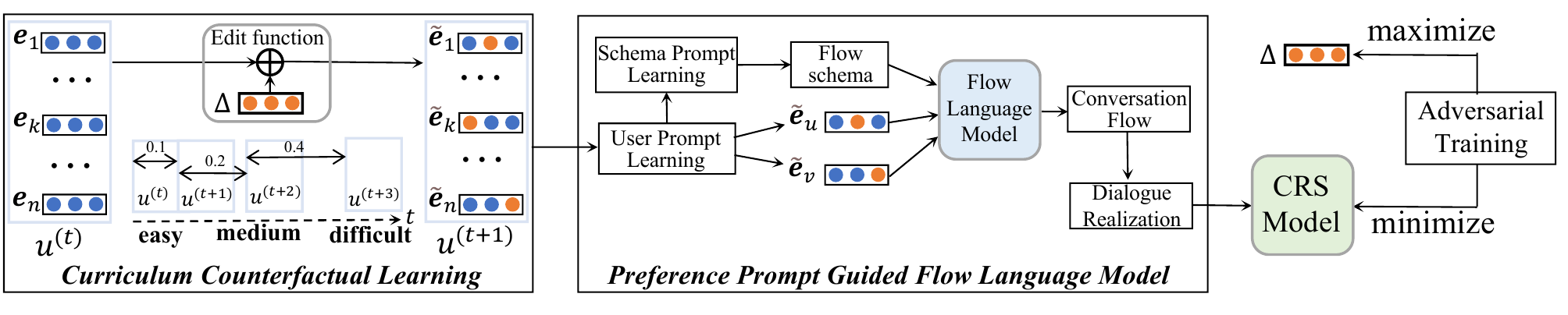}
  \caption{
    The overview of our approach CFCRS. 
    We first adopt curriculum counterfactual learning to augment the user preference at the representation level, and then use the flow language model guided by user and schema prompts to generate conversation flows, which are then realized into dialogues. 
    The edit function and CRS model are optimized with adversarial training to improve both the quality of the augmented data and the recommendation performance.
    }
  \label{model}
\end{figure*}

\section{Approach}

In this section, we present the proposed \textbf{C}ounter\textbf{F}actual data simulation approach for \textbf{CRS}, named \textbf{\framework}, for alleviating the issue of data scarcity in CRSs, which is depicted in \figurename~\ref{model}.

\subsection{Overview of Our Approach}

\paratitle{Task Formulation.} 
Conversational recommender systems~(CRSs) aim to provide accurate item recommendation services through multi-turn natural language conversations. 
At each turn, the system either makes recommendations or chats with the user for preference elicitation.
Such a process ends when the user accepts the recommended items or leaves.
Formally, at the $(j+1)$-th turn, given the dialogue history $C_{j}=\{s_k\}_{k=1}^{j}$ consisting of $j$-turn utterances and the item set $\mathcal{I}$, the system should (1) select a set of candidate items $\mathcal{I}_j$ from the entire item set $\mathcal{I}$ to recommend, and (2) generate the response utterance $s_{j+1}$ to the user. 
Besides, knowledge graph~\cite{chen2019towards, zhou2022c2} as an important auxiliary resource is usually available, denoted by $\mathcal{G}$.  
Typically, a CRS consists of the recommender module (parameterized by $\Theta_R$) and the conversation module (parameterized by $\Theta_C$), which are responsible for the recommendation and response generation tasks, respectively.  

\paratitle{General Model Learning}. 
Formally, let $\mathcal{D} = \{ \langle C_j, r_j, i_j \rangle \}$ denote the set of training samples, where $C_j$ is the dialogue history, $r_j$ is the ground-truth response, and $i_j$ is the recommended item for the $j$-th training sample.  
The optimization objectives for the two modules can be denoted as follows:
\begin{align}
    L_{\Theta_R}(\mathcal{D}) & = -\sum_{ \langle C_j, i_j \rangle \in D} \log g(i_j | C_j; \Theta_R), \\
    L_{\Theta_C}(\mathcal{D}) & = -\sum_{\langle C_j, r_j \rangle \in D} \log h(r_j | C_j; \Theta_C),
\end{align}
where $g(\cdot)$ and $h(\cdot)$ are the recommender and conversation modules, respectively. 
In the literature, existing CRSs mainly focus on designing various models or architectures to implement the two modules. 
For the recommendation module, it can be implemented with collaborative filtering~\cite{li2018towards}, GNN~\cite{zhou2020improving}, or Transformer~\cite{zou2022improving} models.
For the conversation module, it can be implemented with the vanilla Transformer~\cite{chen2019towards} or PLM~\cite{wang2022towards}. 
These approaches rely on high-quality CRS datasets to train the underlying models, which are often limited in size.
To address this limitation, we propose to simulate high-quality data for conversation recommendation, which can be generally applied to various CRSs.

\paratitle{Counterfactual Learning for Dialogue Simulation.}
Our approach is inspired by the recent progress on \emph{counterfactual data augmentation}~\cite{mouli2022bias},  which incorporates counterfactual learning for augmenting the limited data.
In our setting, the essence of our approach is to answer the core question:
``\emph{What the dialogues would be if we intervene on the observed user preference?}''.   
As the basis of our approach, we design a recommendation-oriented dialogue simulator (short as \emph{recommendation dialogue simulator}) that can generate \emph{reasonable, coherent} conversations tailored for two target users (Section~\ref{sec:simulator}). 
Our recommendation dialogue simulator adopts a multi-stage generation process guided by the learned user preference and dialogue sketch: \emph{flow schema} $\rightarrow$ \emph{conversation flow} $\rightarrow$ \emph{dialogue realization}. 
Based on the simulator, we construct the data augmentation via counterfactual learning (Section~\ref{sec-CL}), which performs the intervention at the representations of the interacted entities of a target user. 
Further, we design an adversarial training method with a curriculum schedule that can gradually optimize the edit function towards an improved recommendation capacity of CRSs.
In what follows, we introduce the two parts in detail.

\subsection{Recommendation Dialogue Simulator}
\label{sec:simulator}

The goal of the recommendation dialogue simulator is to generate recommendation-oriented conversation data, so as to improve the performance of existing CRSs. 
Typically, it is difficult to create \emph{fluent, coherent} conversation data, since it needs to simulate the free interaction for information seeking between two real users via chit-chat.  
As our solution, we develop a multi-stage generation process that first generates the \emph{conversation flow} according to the predicted \emph{flow schema} and then realizes the dialogue based on the generated flow. 
In our approach, we first introduce the basic concepts of conversation flow and flow schema.

\subsubsection{Conversation Flow and Flow Schema}
The conversation flow explicitly traces the key elements (\ie entities) of the information-seeking process. 
For example, given a two-turn conversation:

[Seeker]: \emph{I love all kinds of \underline{comedy} movies.} 

[Recommender]: \emph{Have you seen \underline{21 Jump Street?}} 

[Seeker]: \emph{Yes, I love this film because \underline{Jonah Hill} is in it.} 

[Recommender]: \emph{Try another \underline{comedy} movie with him, \underline{Superbad}.} 

\noindent we can derive a conversation flow: \emph{comedy $\rightarrow$  21 Jump Street  $\rightarrow$ Jonah Hill $\rightarrow$   comedy $\rightarrow$  Superbad}. 
Based on such a flow, we can further generalize it into a flow schema: \emph{genre $\rightarrow$  item $\rightarrow$ actor $\rightarrow$ genre $\rightarrow$ item}. 

Formally, a conversation flow $f_{u,v}$ between two users $u$ and $v$ is characterized as a sequence of mentioned entities in a conversation arranged in the occurrence order, denoted by $f_{u,v} = \langle e_1, \cdots,  e_j, \cdots, e_n \rangle$, and the corresponding flow schema (with an equal length to the flow) is characterized as a sequence of type tokens, denoted by $s_{u,v} = \langle t_1, \cdots,  t_j, \cdots, t_n \rangle$, where $e_i$ is a mentioned entity from a knowledge graph~(KG) $\mathcal{E}$ and $t_i=\text{Type}(e_i)$ indicating the type of $e_i$.  
As we can see, conversation flow and flow schema are useful to generate concrete conversation content by capturing the entity preferences of users and establishing the dialogue sketch.

\subsubsection{Preference Prompt Guided Flow Language Model}
\label{sec:prompt}

In our approach, we design a flow language model~(FLM) that is parameterized by $\Theta_F$ based on the \emph{preference prompts} for generating the conversation flow. 
Specifically, to generate a conversation flow, we first sample two target users $u$ and $v$ as the seeker and recommender, respectively, and then employ the two users to predict a flow schema $\hat{s}_{u,v}$. 
Then, the representations of target users (\ie user prompt) and the predicted schema (\ie schema prompt) are taken as the prompts to the FLM for generating the conversation flow as follows: 
\begin{equation}\label{eq-flow-prediction}
f_{u,v} \leftarrow \text{FLM~}\bigg(\bigg[\underbrace{\bm{e}_u, ~\bm{e}_v}_{\text{user prompt}}, ~\underbrace{\{ \bm{t}_{j} \}_{j=1}^n}_{\text{schema prompt}} \bigg]; ~~~\Theta_F\bigg). 
\end{equation}
Next, we discuss how to derive the two parts of prompts. 

\paratitle{User Prompt Learning}.
Since the simulated dialogue occurs between the two target users, it is important to consider their preferences for generating the conversation flow. 
To capture the user preference, in recommender systems, it is common to assign each user a unique user ID, and learn the ID embedding based on the interacted items or entities as the user preference representation~\cite{kang2018self,DBLP:journals/corr/HidasiKBT15}. 
However, in our simulation setting, we would like to generate more diverse user representations that are not limited to the real users in the CRS datasets.
For this purpose, we do not explicitly maintain a user ID but learn ID-agnostic user representations.
Specifically, following the previous work on CRSs~\cite{zhou2020improving,zhang2022variational}, we assume that a KG is available and extend this KG by attaching user nodes to their interacted entity nodes to compose a new heterogeneous knowledge graph~(HKG), denoted as $\mathcal{G}$. 
To capture relational semantics between entities, we utilize R-GCN~\cite{schlichtkrull2018modeling} to learn entity representations on the HKG. 
Formally, let $n$ denote a node placeholder for the HKG, associated with an embedding vector $\bm{e}_n \in \mathbb{R}^{d_E}$ derived from R-GCN, where $d_E$ denotes the embedding size.
We utilize the self-attention mechanism to aggregate entity embeddings as the preference representation of the user $u$:
\begin{align}\label{eq-u-prompt}
    \bm{e}_u    & = \mathbf{E}_u \cdot \bm{\alpha},                                                   \\\notag
    \bm{\alpha} & = \text{softmax}( \bm{b}^\top \cdot \text{tanh}(\mathbf{W}_{\alpha} \mathbf{E}_u)),
\end{align}
where $ \mathbf{E}_u$ is the matrix consisting of the embeddings of all the interacted entities of user $u$, $\bm{\alpha}$ is the attention weight vector reflecting the importance of each entity, and $\mathbf{W}_{\alpha} $ and $\bm{b}$ are trainable parameters. 
A major advantage of this representation method is that it can be easily adapted to new users by modeling the entity preference based on the associated interaction records. 

\paratitle{Schema Prompt Learning}.
In order to produce \textit{fluent, coherent} conversations, we employ frequent flow schema to structure the conversation. 
Although the conversation flows can be very diverse, the frequent flow schemas for a CRS corpus are usually limited. 
Thus, we consider employing real CRS datasets to construct flow schemas with frequent pattern mining algorithms~\cite{han2007frequent}, and obtain a set of frequent flow schemas, denoted as $\mathcal{S}$. 
Then, the schema prediction task is cast as a classification problem over the schema set $\mathcal{S}$ based on user preference.
Formally, we compute the prediction probability for a flow schema as follows:
\begin{equation}
    \label{eq:schema}
    \text{Pr}(s_{u,v} | u, v) = \text{softmax}\big(\text{MLP}([\bm{e}_u, \bm{e}_v])\big),
\end{equation}
where $u$ and $v$ are the two target users involved in the dialogue, and $s_{u,v}$ is the flow schema. 
In practice, first, we select the most probable flow schema according to Eq.~\eqref{eq:schema}. 
Then, we obtain the corresponding type embeddings by decomposing the predicted schema into type tokens $\{ \bm{t}_j \}_{j=1}^n$, where each type token embedding $\bm{t}_j$ is obtained by looking up the type embedding table. 

\paratitle{Flow Language Model Pre-Training}.
To model the conversation flow, we construct an FLM based on Transformer, which utilizes the encoder-decoder architecture.
Following Eq.~\eqref{eq-flow-prediction}, the encoder takes the learned user prompt (\ie $\bm{e}_u$ and $\bm{e}_v$) and schema prompt (\ie $\{ \bm{t}_j \}_{j=1}^n$) as input, and the decoder generates the conversation flow in an autoregressive manner based on the prompt. 
Formally, let $\bm{e}_{j}$ be the embedding of a token $e_{j}$ (an entity) in the conversation flow $f_{u,v}$, and the probability of the flow to be generated is formulated as:
\begin{align}
    \text{Pr}(f_{u,v}) &= \sum_{j=1}^n \text{Pr}(e_j | e_0, \dots, e_{j-1}) \\
                       &= \sum_{j=1}^n \text{softmax} \big( \mathbf{W} [\bm{e}_{j}; \bm{z}_{u,v}] + \bm{b} \big) \nonumber
\end{align}
where $\bm{z}_{u,v}$ is the encoding of prompt, $\mathbf{W}$ and $\bm{b}$ are trainable parameters. 
To pre-train the FLM, we require large amounts of conversation flows from various user pairs.
Since the original CRS dataset is usually limited in conversation size, we consider generating pseudo conversation flows for pre-training. 
The basic procedure consists of three major steps:
(i) we first randomly sample one flow schema from the frequent flow schema set $\mathcal{S}$;
(ii) then, we sample an entity sequence from the constructed HKG $\mathcal{G}$ as the conversation flow according to the schema;
(iii) finally, we randomly divide the entities in the sequence into two groups, which correspond to the entity preference of two users involving the conversation.
When a schema cannot lead to a reachable entity path, we continue to sample another schema. 

\subsubsection{Dialogue Realization}

With the pre-trained FLM and entity preference of users, we can generate the corresponding conversation flows at a large scale.
Next, we realize the generated conversation flows into recommendation dialogues.

Here, we adopt a simple template-based approach for dialogue realization.
Specifically, we first collect the templates from observed dialogues by delexicalization, \ie substituting the mentioned entities with placeholders, \eg ``\emph{<genre>}'' for ``comedy''. 
Then, the entities in conversation flows can be sequentially filled into these templates as new recommendation dialogues.
For instance, an utterance ``\emph{I am in a mood for something \underline{scary}}'' would be converted to the template ``\emph{I am in a mood for something <genre>}''.
Since the focus of this work is to enhance the recommendation ability of CRSs instead of the general chit-chat ability, we do not adopt pre-trained dialogue models (\eg DialogGPT~\cite{zhang2020dialogpt}) to realize these utterances. 
Our proposed method is simple yet effective to ensure fluency in language and faithfulness in conversation flow. 

After building the recommendation dialogue simulator, we can employ it to generate simulated data with user preference representations as input (as will be used in Section~\ref{sec-CL}).  

\subsection{Curriculum Counterfactual Learning}
\label{sec-CL}

Although the above recommendation dialogue simulator can effectively enlarge the dataset by simulation, it is still limited to the actual users in existing datasets. 
In this part, we introduce a curriculum counterfactual learning approach that can learn to generate diverse data by augmenting the user preference representations. 

\subsubsection{Counterfactual User Preference Augmentation}

Recall that in Section~\ref{sec:prompt}, we utilize a self-attentive mechanism to learn the user preference representation based on the interacted entities with Eq.~\eqref{eq-u-prompt}.  
Formally, the set of interacted entities of user $u$ is denoted as $\mathcal{E}_u = \{e_1, \dots, e_i, \dots, e_n\}$, and their embeddings are also aggregated as a matrix $\mathbf{E}_u = [ \bm{e}_1, \dots, \bm{e}_i, \dots, \bm{e}_n ]$. 

In existing work~\cite{liu2021augmenting,qiu2021memory}, discrete~\cite{liu2021augmenting} and continuous~\cite{qiu2021memory} item-level edits have been explored for user data augmentation. 
To combine the merits of both approaches, we consider an entity-level edit to augment new user preferences. 
Specifically, we revise one entity embedding at each time with a specially designed edit function $f(\cdot)$, in which the entity selection is \emph{discrete} and the embedding revision is \emph{continuous}.    
Such a way can generate diverse user preferences while gradually incorporating controllable revisions.
To instantiate the edit function, a number of model choices can be considered, \eg neural networks.
However, we empirically find that it is difficult to optimize such an edit function, due to the lack of supervision signals in real datasets.
Thus, we consider a simple yet effective edit function that directly adds a small disturbance vector $\bm{\Delta}_i$: 
\begin{equation}\label{eq:edit}
f(\bm{e}_i) = \bm{e}_i + \bm{\Delta}_i, 
\end{equation}
We denote all the disturbance vectors as $\Theta_E$. 

According to~\cite{goyal2019counterfactual}, such a simplified edit function is easier to learn and interpret, since samples near the decision boundary are usually discriminative in revealing the underlying data patterns.  
For each user $u$, we can perform the edit $k$ times, so as to produce $k$ different augmentations, each editing one specific entity embedding in $\mathbf{E}_u$ to generate the augmented user preference $\tilde{\bm{e}}_u$ (originally $\bm{e}_u$).   



\subsubsection{Adversarial Learning with Curriculum Schedule}

As our edit function can be learned in a differentiable manner, we propose to use adversarial learning to enhance the informativeness of the augmented user preference representations. 
Intuitively, a more informative training instance tends to cause \emph{a larger loss} in the recommendation accuracy, as such a user preference has not been well captured by the \emph{current model}.  
Taking an adversarial learning perspective, the counterfactual edit function (parameterized by $\Theta_E$) aims to maximize the loss of the recommender module (parameterized by $\Theta_R$), while the recommender module aims to minimize its loss on the simulated data.  
In addition, we perform adversarial learning with a curriculum schedule to stabilize the optimization.

\paratitle{Adversarial Training}. 
Let $\pi_{\Theta_E}(C | \tilde{\bm{e}}_u, \tilde{\bm{e}}_v)$ denote the recommendation dialogue simulator, which returns the probability of generating a dialogue $C$ given the edited embeddings of two users $\tilde{\bm{e}}_u$ and $\tilde{\bm{e}}_v$ by $\Theta_E$.  
The learning objective can be formulated as follows: 
\begin{equation}
    \label{eq:adversarial}
    J^{\Theta_R^*, \Theta_E^*} = \min_{\Theta_R} \max_{\Theta_E} \mathbb{E}_{C \sim \pi_{\Theta_E} (\cdot | \tilde{\bm{e}}_u, \tilde{\bm{e}}_v )} [  L_{\Theta_R}(C) - \lambda \cdot \parallel \Theta_E \parallel_2^2 ],
\end{equation}
where $\Theta_E=\{ \bm{\Delta}_u, \bm{\Delta}_v\}$ is the edit vectors for users $u$ and $v$ (Eq.~\eqref{eq:edit}),  $L_{\Theta_R}(C)$ is the loss of the recommendation module for the generated data $C$, and $\lambda$ is the regularization weight for $\Theta_E$.
Note that Eq.~\eqref{eq:adversarial} presents the optimization objective for a pair of users, which can be easily extended to all user pairs. To optimize the above objective (with two groups of parameters $\Theta_R$ and $\Theta_E$), we can alternatively optimize each group of parameters by keeping the other group fixed.  It is relatively straightforward to train the parameters of the recommender module (\ie $\Theta_R$) by a standard recommendation loss (\eg cross-entropy loss~\cite{ren2022variational}). 
However, it is infeasible to directly optimize the edit vectors in an end-to-end way, since it involves the generation of discrete conversation data. 
To tackle this issue, we adopt the classic REINFORCE algorithm~\cite{williams1992simple} to update the parameters as follows: 
\begin{equation}\label{eq:REINFORCE}
    \Theta_E = \Theta_E + \alpha \bigg( \sum_{t=1}^{T} L_{\Theta_R}(C_t) \nabla_{ \Theta_E } \log\big(\pi_{ \Theta_E}(C_t | \tilde{\bm{e}}_u, \tilde{\bm{e}}_v )\big) - 2 \lambda \cdot \Theta_E\bigg),
\end{equation}
where $\alpha$ is the learning rate and $T$ conversations are sampled.

\paratitle{Curriculum Arrangement.}
In order to keep the training stable, we consider a curriculum learning approach that gradually increases the augmentation level: small variations are encouraged at the beginning of training, while larger variations can be gradually applied to enhance the model capacity. 
As shown in Eq.~\eqref{eq:adversarial}, we incorporate a controlling weight $\lambda$ on $\Theta_E$. 
To simulate counterfactual data in an \emph{easy-to-difficult} process, we dynamically tune the augmentation level in each iteration. 
Specifically, we apply an annealing mechanism to regularize $\Theta_E$ with a shrinking $\lambda$:
\begin{equation}  \label{eq:curriculum}
\lambda = \rho \times \delta^{(k)},
\end{equation}
where $\rho$ is the initial weight, $\delta$ is the decay ratio, and $k$ is the current course. 
In this way, as the course gets more difficult, \ie the augmentation level increases, the CRS model can continually learn from diverse and informative training samples to improve its performance.

\subsection{Parameter Learning}

\begin{algorithm}[t]
\small
    \caption{The training algorithm of our framework.}
    \label{algorithm}
    \LinesNumbered
    \KwIn{
        The conversational recommendation dataset $\mathcal{D}$, HKG $\mathcal{G}$
    }
    \KwOut{Parameters of the recommender module $\Theta_{R}$ and conversation module $\Theta_{C}$ in CRSs.}
 
    Pre-train the parameters of the recommendation dialogue simulator $\Theta_{S}$ with the union of real and pseudo data sampled from $\mathcal{G}$. \\
    Pre-train the parameters of the recommender module $\Theta_{R}$ using the real dataset $\mathcal{D}$. \\
    \For {$k = 1 \to N$} {
        Set the regularization parameter $\lambda$ according to curriculum arrangement using Eq.~\eqref{eq:curriculum}. \\
        Optimize the parameters of edit function $\Theta_{E}$ by maximizing the loss of the recommender module using Eq.~\eqref{eq:REINFORCE} and derive new user preference $\tilde{\bm{e}}$. \\
        Use new user preference $\tilde{\bm{e}}$ to simulate data $C$ with the recommendation dialogue simulator by Eq.~\eqref{eq-flow-prediction} and Eq.~\eqref{eq-u-prompt}. \\
        Optimize the parameters of the recommender module $\Theta_{R}$ by minimizing its loss on simulated data $C$. \\
    }
    Optimize the conversation module $\Theta_{C}$ with the augmented data. \\
    \Return $\Theta_{R}$ and $\Theta_{C}$.
\end{algorithm}

The parameters of our framework consist of four groups, namely the recommendation dialogue simulator $\Theta_S$, the counterfactual edit function $\Theta_E$, and the recommender module $\Theta_R$ and conversation modules $\Theta_C$ of the target CRS model.
Algorithm~\ref{algorithm} presents the training algorithm of our framework.

First of all, we pre-train the parameters of the recommendation dialogue simulator $\Theta_S$ with the union of real and pseudo conversation flow data using the cross-entropy loss.
After pre-training, parameters $\Theta_S$ are fixed.
Then, we perform curriculum counterfactual learning to augment new data for CRS learning.
In each iteration, the parameters of the counterfactual edit function $\Theta_E$ and the recommender module of the target CRS model $\Theta_R$ are optimized via adversarial learning using Eq.~\eqref{eq:adversarial}.
Specifically, we first learn the counterfactual edit function to maximize the loss of the recommender module, and then optimize the recommender module to minimize its loss on the simulated data.
After the curriculum learning schedule, we optimize the parameters of the conversation module $\Theta_C$ with the union of simulated and real data.

\section{Experiment}
In this section, we first set up the experiments, then report the results and give a detailed analysis.

\subsection{Experimental Setup}

\paratitle{Datasets.}
To verify the effectiveness of our approach, we conduct experiments on two widely used English CRS datasets, \ie \textsc{ReDial}~\cite{li2018towards} and \textsc{INSPIRED}~\cite{hayati2020inspired}.
The \textsc{ReDial} dataset is an English CRS dataset about movie recommendations, and is constructed through crowdsourcing workers on Amazon Mechanical Turk (AMT).
Similar to \textsc{ReDial}, the \textsc{INSPIRED} dataset is also an English CRS dataset about movie recommendations, but with a much smaller size.
The statistics of both datasets are summarized in \tablename~\ref{tab:datasets}.

\begin{table}[t]
    \centering
    \caption{Statistics of the datasets after preprocessing.}
    \small
    \label{tab:datasets}
    \begin{tabular}{crrr}
        \toprule
        \textbf{Dataset} & \textbf{\#Dialogues} & \textbf{\#Utterances} & \textbf{\#Items} \\
        \midrule
        INSPIRED         & 1,001                & 35,811                & 1,783            \\
        ReDial           & 10,006               & 182,150               & 51,699           \\
        \bottomrule
    \end{tabular}
\end{table}

\begin{table*}[t]
\centering
\caption{
    Results on the recommendation task. The best methods in each group are marked in bold.
    Numbers marked with * indicate that the improvement is statistically significant compared with the baseline (t-test with p-value < 0.05).
}
\label{tab:rec}
\resizebox{\textwidth}{!}{%
\begin{tabular}{l|cccccc|cccccc}
\toprule
Datasets     & \multicolumn{6}{c}{ReDial}                    & \multicolumn{6}{c}{INSPIRED}                  \\
\midrule
Models & Recall@10 & Recall@50 & MRR@10 & MRR@50 & NDCG@10 & NDCG@50 & Recall@10 & Recall@50 & MRR@10 & MRR@50 & NDCG@10 & NDCG@50 \\
\midrule
ReDial       & 0.129 & 0.287 & 0.003 & 0.004 & 0.005 & 0.011 & 0.117 & 0.285 & 0.004 & 0.003 & 0.005 & 0.012 \\
BERT         & 0.156 & 0.357 & 0.055 & 0.063 & 0.079 & 0.121 & 0.179 & 0.328 & 0.067 & 0.085 & 0.098 & 0.133 \\
GPT-2        & 0.147 & 0.327 & 0.051 & 0.056 & 0.071 & 0.107 & 0.112 & 0.278 & 0.063 & 0.076 & 0.089 & 0.128 \\
DialoGPT     & 0.173 & 0.361 & 0.062 & 0.068 & 0.089 & 0.135 & 0.125 & 0.247 & 0.059 & 0.081 & 0.092 & 0.120 \\
\midrule
KBRD         & 0.170 & 0.366 & 0.063 & 0.072 & 0.088 & 0.131 & 0.210 & 0.390 & 0.112 & 0.118 & 0.135 & 0.172 \\
KBRD-EDA     & 0.174 & 0.371 & 0.068 & 0.077 & 0.093 & 0.136 & 0.180 & 0.364 & 0.088 & 0.094 & 0.109 & 0.146 \\
KBRD-mixup   & 0.189 & 0.390 & 0.072 & 0.081 & 0.099 & 0.144 & 0.210 & 0.390 & 0.104 & 0.113 & 0.122 & 0.165 \\
KBRD-CFCRS      & \textbf{0.206}* & \textbf{0.408}* & \textbf{0.084}* & \textbf{0.093}* & \textbf{0.109}* & \textbf{0.156}* & \textbf{0.226}* & \textbf{0.426}* & \textbf{0.123}* & \textbf{0.129}* & \textbf{0.145}* & \textbf{0.188}* \\
\midrule
BARCOR       & 0.169 & 0.374 & 0.063 & 0.073 & 0.088 & 0.133 & 0.185 & 0.339 & 0.080 & 0.087 & 0.104 & 0.137 \\
BARCOR-EDA   & 0.179 & 0.395 & 0.067 & 0.077 & 0.093 & 0.140 & 0.210 & 0.390 & 0.102 & 0.109 & 0.127 & 0.166 \\
BARCOR-mixup & 0.169 & 0.363 & 0.061 & 0.070 & 0.086 & 0.129 & 0.139 & 0.344 & 0.070 & 0.081 & 0.086 & 0.132 \\
BARCOR-CFCRS    & \textbf{0.198}* & \textbf{0.406}* & \textbf{0.079}* & \textbf{0.088}* & \textbf{0.107}* & \textbf{0.151}* & \textbf{0.246}* & \textbf{0.421}* & \textbf{0.114}* & \textbf{0.122}* & \textbf{0.145}* & \textbf{0.183}* \\
\midrule
UniCRS       & 0.217 & 0.428 & 0.088 & 0.096 & 0.118 & 0.163 & 0.272 & 0.441 & 0.156 & 0.164 & 0.184 & 0.224 \\
UniCRS-EDA   & 0.167 & 0.357 & 0.068 & 0.077 & 0.091 & 0.133 & 0.295 & 0.451 & 0.132 & 0.165 & 0.186 & 0.220 \\
UniCRS-mixup & 0.206 & 0.394 & 0.073 & 0.088 & 0.116 & 0.158 & 0.246 & 0.426 & 0.153 & 0.164 & 0.182 & 0.219 \\
UniCRS-CFCRS    & \textbf{0.231}* & \textbf{0.444}* & \textbf{0.096} & \textbf{0.111}* & \textbf{0.129}* & \textbf{0.175}* & \textbf{0.308}* & \textbf{0.466}* & \textbf{0.168}* & \textbf{0.176}* & \textbf{0.204}* & \textbf{0.242}* \\
\bottomrule
\end{tabular}%
}
\end{table*}

\paratitle{Baselines.}
Here we consider two major tasks for CRS evaluation, namely recommendation and conversation.
For comparison, we select several representative methods (including both CRS models and adapted PLMs) tailored to each task.



\textbullet~{\textit{BERT}}~\cite{devlin2019bert}:
It is a bidirectional PLM pre-trained via the masked language modeling task on a large-scale general corpus.
We utilize the representation of the $[CLS]$ token for recommendation.

\textbullet~{\textit{GPT-2}}~\cite{radford2019language}:
It is an autoregressive PLM pre-trained via the language modeling task on large-scale general corpora. 
We concatenate the utterances in the conversation history as inputs, and take the generated text for response while using the representation of the last token for recommendation.

\textbullet~{\textit{DialoGPT}}~\cite{zhang2020dialogpt}:
It continues to pre-train GPT-2 on large-scale dialogue corpora.
We use it in the same way as GPT-2.

\textbullet~{\textit{ReDial}}~\cite{li2018towards}:
It is proposed along with the \textsc{ReDial} dataset, which includes a conversation module based on HRED~\cite{subramanian2018learning} and a recommendation module based on a denoising auto-encoder~\cite{he2017neural}.

\textbullet~{\textit{KBRD}}~\cite{chen2019towards}:
It introduces DBpedia to enhance the semantics of entities mentioned in the dialogues.

\textbullet~{\textit{BARCOR}}~\cite{wang2022barcor}:
It proposes a unified framework based on BART, which tackles two tasks with a single model.

\textbullet~{\textit{UniCRS}}~\cite{wang2022towards}:
It designs knowledge-enhanced prompts based on DialoGPT to fulfill both tasks in a unified approach.

Among these baselines, BERT, GPT-2, and DialoGPT are PLMs, where BERT and GPT-2 is pre-trained on general corpora while DialoGPT is pre-trained on dialogue corpora.
We fine-tune these PLMs to encode the dialogue and generate items to recommend and utterances to respond to.
ReDial, KBRD, BARCOR, and UniCRS are CRS methods, where ReDial and KBRD use mentioned entities for recommendation, BARCOR utilizes dialogue texts for recommendation, and UniCRS makes use of both entities and texts for recommendation.
To verify the generality of our framework, we apply it to KBRD, BARCOR, and UniCRS.
To demonstrate the effectiveness of our framework, we compare it with several representative data augmentation methods.

\textbullet~{\textit{EDA}}~\cite{DBLP:conf/emnlp/WeiZ19}:
It augments new examples by randomly performing edit operations, \ie replacement, insertion, swap, and deletion.

\textbullet~{\textit{Mixup}}~\cite{zhang2017mixup}:
It augments new examples in the continuous latent space by linear interpolations of input representations and labels of two random examples.


\paratitle{Evaluation Metrics.}
Following existing work~\cite{li2018towards,chen2019towards}, we adopt different metrics to evaluate the recommendation and conversation tasks separately.
For the recommendation task, following~\cite{chen2019towards,zhou2020towards}, we use Recall@$k$, MRR@$k$, and NDCG@$k$ ($k$=10,50).
For the conversation task, following~\cite{zhou2020improving,zhou2022c2}, we adopt Distinct-$n$ ($n$=2,3,4) to evaluate the diversity of the generated responses.
Besides, following KGSF~\cite{zhou2020improving}, we invite three annotators to score the generated responses from two aspects, namely \emph{Fluency} and \emph{Informativeness}. 
The range of scores is 0 to 2.
For all the above metrics, we calculate and report the average scores on all test examples.

\paratitle{Implementation Details.}
We implement all the baseline models based on the open-source toolkit CRSLab~\cite{zhou2021crslab}~\footnote{https://github.com/RUCAIBox/CRSLab}, which contains comprehensive CRS models and benchmark datasets.
For the recommendation dialogue simulator, we adopt a Transformer with 12-layer encoders and decoders as the FLM, and its hidden size and embedding size are 768.
To be consistent with the FLM, the hidden size of the user prompt and schema prompt is also 768.
In curriculum counterfactual learning, the maximum training iterations are set to 20, and we adopt the early stopping strategy.
The initial value of the regularization weight and the decay ratio is tuned in the range of $[10^{-1}, 10^{-2}, 10^{-3}]$ and $[0.9, 0.8, 0.7]$ for different models.
We use AdamW~\cite{loshchilov2018decoupled} with the default parameter setting to optimize the parameters in our framework.
The learning rate is mostly set to $1e^{-4}$ and tuned in the range of $[5e^{-5}, 1e^{-4}, 5e^{-4}, 1e^{-3}]$.

\subsection{Evaluation on Recommendation Task}

In this part, we conduct experiments to evaluate the effectiveness of our model on the recommendation task.

\paratitle{Automatic Evaluation.}
Table~\ref{tab:rec} shows the performance of different methods on the recommendation task.
First, we can see that KBRD, BARCOR, and UniCRS mostly outperform the other baselines in all metrics.
The three methods all incorporate external KGs to enrich the information of mentioned entities in the conversation context, which can effectively alleviate the data scarcity problem and better capture user intents and preferences.
Among the three methods, UniCRS performs the best in all metrics.
UniCRS utilizes knowledge-enhanced prompts to guide the PLM, and incorporates a pre-training task to improve the quality of prompts.
Such a way can effectively endow the PLM with entity knowledge for better performance.

Second, for the two data augmentation baselines, we observe that most of the time they both improve the performance of the three CRS methods.
It indicates the effectiveness of data augmentation strategies in the CRS task, since the training data is not sufficient.
However, we can see that the improvement is not stable, and even causes performance degradation for UniCRS on the \textsc{ReDial} dataset.
A possible reason is that the two methods only rely on heuristic rules to modify the original examples for augmenting new ones, which makes it hard to guarantee the quality of the augmented data and may even produce abnormal conversations.

Finally, we can see that our model can improve the performance of the three CRS methods by a large margin.
It indicates the effectiveness and generality of our framework.
Furthermore, our approach mostly outperforms the two data augmentation baselines significantly.
In our approach, we use a counterfactual data simulation approach, which includes a pre-trained FLM to guarantee the coherence of the conversation flow and adversarial training to enhance the informativeness of simulated data.
Besides, we utilize the curriculum learning strategy to gradually optimize CRS models using examples with different augmentation levels, which further improves the stability of the training process.

\begin{table}[t]
\centering
\caption{
    Ablation analysis on the recommendation task. ``FLM'' denotes the flow language model and ``Template'' denotes template-based dialogue realization. ``-'' denotes removing the corresponding component.
}
\label{tab:rec-ab}
\resizebox{\columnwidth}{!}{%
\begin{tabular}{l|cc|cc}
\toprule
Datasets     & \multicolumn{2}{c}{ReDial} & \multicolumn{2}{c}{INSPIRED} \\
\midrule
Metrics      & Recall@10            & Recall@50             & Recall@10             & Recall@50         \\
\midrule
BARCOR       & 0.169                & 0.374                 & 0.185                 & 0.339             \\
+CFCRS       & \textbf{0.198}       & \textbf{0.406}        & \textbf{0.246}        & \textbf{0.421}    \\
\midrule
-Curriculum  & 0.187                & 0.399                 &  0.190                & 0.405             \\
-Adversarial & 0.184                & 0.389                 &  0.225                & 0.395             \\
-FLM         & 0.181                & 0.385                 &  0.211                & 0.390             \\
-Frequent Schema & 0.186                & 0.394                 &  0.246                & 0.407             \\
-Template    & 0.183                & 0.382                 &  0.174                & 0.385             \\
\bottomrule
\end{tabular}%
}
\end{table}

\paratitle{Ablation Study.}
Our approach incorporates several important components to improve the quality of the augmented data.
To verify the effectiveness of each component, we conduct the ablation study on BARCOR using the \textsc{ReDial} and \textsc{INSPIRED} datasets.
We report the results of Recall@10 and Recall@50.
We consider removing the curriculum schedule, the adversarial training objective, the FLM, the frequent flow schemas, and the template-based dialogue realization, respectively.

The results are shown in Figure~\ref{tab:rec-ab}.
We can see that removing any component would lead to performance degradation.
It indicates that all the components in our model are useful to improve the performance of the recommendation task.
Among them, performance decreases the most after removing the template-based dialogue realization.
It indicates that the template-based dialogue realization is important in our approach, since it can ensure fluency in language and faithfulness in conversation flow without introducing noise to the simulated data, which is beneficial for the improvement of the recommendation ability of CRSs.

\subsection{Evaluation on Conversation Task}
In this part, we conduct experiments to verify the effectiveness of our model on the conversation task.

\begin{table}[t]
\centering
\caption{
    Automatic evaluation results on the conversation task.
    We abbreviate Distinct-2,3,4 as Dist-2,3,4. 
    The best methods in each group are marked in bold.
    Numbers marked with * indicate that the improvement is statistically significant compared with the baseline (t-test with p-value < 0.05).
}
\label{tab:conv}
\resizebox{\columnwidth}{!}{%
\begin{tabular}{l|ccc|ccc}
\toprule
Datasets     & \multicolumn{3}{c}{ReDial} & \multicolumn{3}{c}{INSPIRED} \\
\midrule
Models       & Dist-2  & Dist-3  & Dist-4 & Dist-2   & Dist-3  & Dist-4  \\
\midrule
ReDial       & 0.023   & 0.236   & 0.228  & 0.153    & 0.255   & 0.397   \\
GPT-2        & 0.354   & 0.486   & 0.441  & 2.347    & 3.691   & 4.568   \\
DialoGPT     & 0.476   & 0.559   & 0.486  & 2.408    & 3.720   & 4.560   \\
\midrule
KBRD         & 0.198   & 0.339   & 0.473  & 0.223    & 0.415   & 0.616   \\
KBRD-EDA     & 0.323   & 0.476   & 0.565  & 0.466    & 0.856   & 1.174   \\
KBRD-mixup   & 0.172   & 0.292   & 0.449  & 0.362    & 0.680   & 0.987   \\
KBRD-CFCRS      & \textbf{0.477}*   & \textbf{0.603}*   & \textbf{0.728}*  & \textbf{0.573}*  & \textbf{1.148}*   & \textbf{1.645}*   \\
\midrule
BARCOR       & 0.404   & 0.540   & 0.654  & 2.923    & 4.172   & 4.992   \\
BARCOR-EDA   & 0.522   & 0.698   & 0.717  & 3.597    & 5.108   & 5.959   \\
BARCOR-mixup & 0.568   & 0.704   & 0.740  & 3.856    & 5.582   & 6.576   \\
BARCOR-CFCRS    & \textbf{0.701}*   & \textbf{0.971}*   & \textbf{0.969}*  & \textbf{4.081}*    & \textbf{5.953}*   & \textbf{6.979}*   \\
\midrule
UniCRS       & 0.351   & 0.631   & 0.897  & 2.809    & 4.530   & 5.555   \\
UniCRS-EDA   & 0.440   & 0.801   & 1.141  & 2.882    & 4.859   & 6.166   \\
UniCRS-mixup & 0.412   & 0.701   & 0.918  & 2.517    & 4.173   & 5.496   \\
UniCRS-CFCRS    & \textbf{0.632}*   & \textbf{1.195}*   & \textbf{1.524}*  & \textbf{4.225}*    & \textbf{6.824}*   & \textbf{8.155}*   \\
\bottomrule
\end{tabular}%
}
\end{table}

\paratitle{Automatic Evaluation.}
We show the evaluation results of automatic metrics about different methods in Table~\ref{tab:conv}.
As we can see, the methods using PLMs (\ie GPT-2, DialoGPT, BARCOR, and UniCRS) mostly achieve better performance than other methods. 
Since PLMs have been pre-trained with generative tasks on large-scale corpora, they can quickly adapt to the CRS task and generate diverse responses after fine-tuning.
Among these methods, UniCRS mostly achieves the best performance.
Since UniCRS is based on DialoGPT, a PLM that has been pre-trained on large-scale dialogue corpora, it is more capable of generating responses.
It also performs semantic fusion and prompt pre-training to inject task-specific knowledge into DialoGPT, helping generate more informative responses.

Besides, we can see that the two data augmentation methods also improve the performance of the three CRS models.
Despite the fact that the improvement is not stable, it can demonstrate that CRS models are hungry for training data.

Finally, our model also consistently boosts the performance of the three CRS models, and significantly outperforms the two data augmentation methods.
It further indicates the effectiveness and generality of our framework among different CRS methods.
Besides, we can see that with the help of our approach, the model KBRD can even outperform PLM-based methods on the \textsc{ReDial} dataset.
It shows that our proposed data augmentation approach suits well with KBRD, and can inspire its potential to generate high-quality responses.

\begin{table}[t]
\centering
\caption{
    Human evaluation results about the conversation task on the \textsc{ReDial} dataset. 
}
\label{tab:conv-human}
\begin{tabular}{lcc}
\toprule
Models       & Fluency & Informativeness \\
\midrule
KBRD         & 0.91    & 0.86            \\
KBRD-EDA     & 1.12    & 1.04            \\
KBRD-mixup   & 1.05    & 0.93            \\
KBRD-CFCRS      & \textbf{1.27}    & \textbf{1.09}            \\
\midrule
BARCOR       & 1.23    & 1.14            \\
BARCOR-EDA   & 1.29    & 1.22            \\
BARCOR-mixup & 1.36    & 1.30            \\
BARCOR-CFCRS    & \textbf{1.47}    & \textbf{1.38}            \\
\midrule
UniCRS       & 1.41    & 1.33            \\
UniCRS-EDA   & 1.57    & 1.49            \\
UniCRS-mixup & 1.48    & 1.41            \\
UniCRS-CFCRS    & \textbf{1.69}    & \textbf{1.60}            \\
\bottomrule
\end{tabular}%
\end{table}

\paratitle{Human Evaluation.}
To provide a more qualified evaluation of the conversation task, we conduct the human evaluation following previous work~\cite{zhou2020improving}. 
We select KBRD, BARCOR, and UniCRS as the backbone, and implement our approach on them.
We invite three annotators to evaluate the \textit{fluency} and \textit{informativeness} of the generated responses from examples from these models, and present the results on the \textsc{ReDial} dataset in Table~\ref{tab:conv-human}.
The average Cohen's kappa between any two annotators is 0.89, which indicates good agreement.

First, we can also see a similar tendency to the automatic metrics: UniCRS $>$ BARCOR $>$ KBRD.
It indicates the effectiveness of UniCRS which incorporates DialoGPT and knowledge-enhanced prompts.
Besides, the two data augmentation methods can consistently improve the quality of the generated response, but their performance order is not stable.
A possible reason is that they rely on heuristic rules for augmentation without considering the target model, which may produce useless examples for specific CRS models.
Although the augmented examples may contain noise, they are still able to alleviate the data-hungry problem of CRS models.
Finally, our approach can consistently outperform these baseline models.
It further demonstrates the effectiveness of our framework, which can augment more high-quality examples to improve the training of CRS models, helping them generate fluent and informative responses.

\subsection{Performance Comparison w.r.t. Different Amount of Training Data}

\begin{figure}[t]
     \centering
     \includegraphics[width=0.49\linewidth]{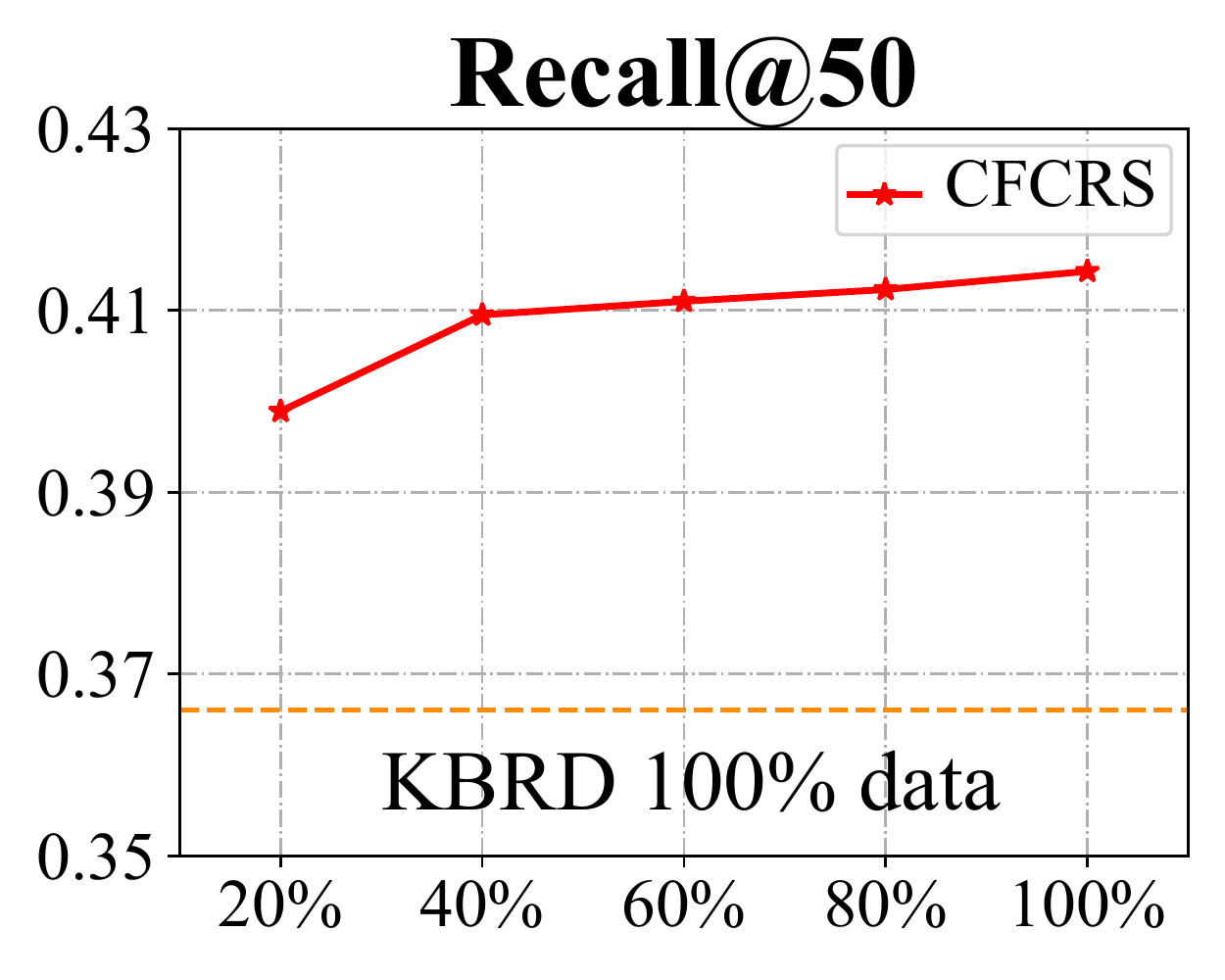}
     \includegraphics[width=0.49\linewidth]{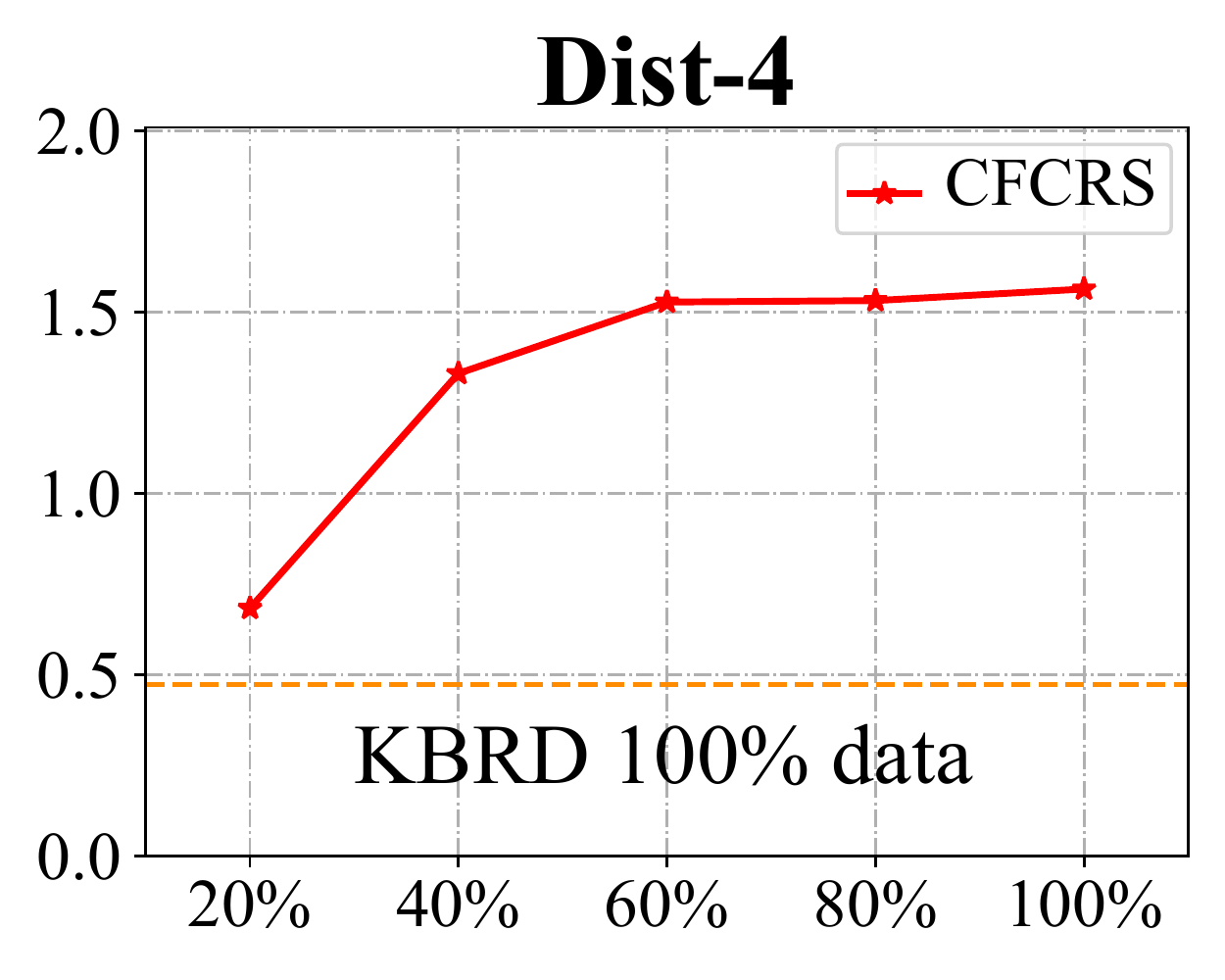}
     \caption{Performance comparison w.r.t. different amounts of training data on the \textsc{ReDial} dataset. We implement our framework on KBRD.}
     \label{fig:few-shot}
\end{figure}

In real-world applications, the data scarcity problem can be more serious and greatly constrain performance.
Since our approach can augment high-quality examples, they can alleviate this problem to some extent.
To validate this, we simulate a data scarcity scenario by sampling different proportions of the training data, \ie 20\%, 40\%, 60\%, 80\%, and 100\%.
We implement our approach on KBRD and report the results of the recommendation and conversation tasks on the \textsc{ReDial} dataset.

Figure~\ref{fig:few-shot} shows the evaluation results in different data scarcity settings.
As we can see, with just 20\% training data, our approach can still achieve comparable performance with KBRD that is trained using 100\% data.
It indicates that our approach can augment high-quality conversations, greatly alleviating the data scarcity problem.
Besides, with less available training data, the performance of our approach is relatively stable.
It also shows the potential of our approach to dealing with the cold-start scenario in real-world applications.

\subsection{Hyper-parameters Analysis}
\begin{figure}[t]
     \centering
     \includegraphics[width=0.49\linewidth]{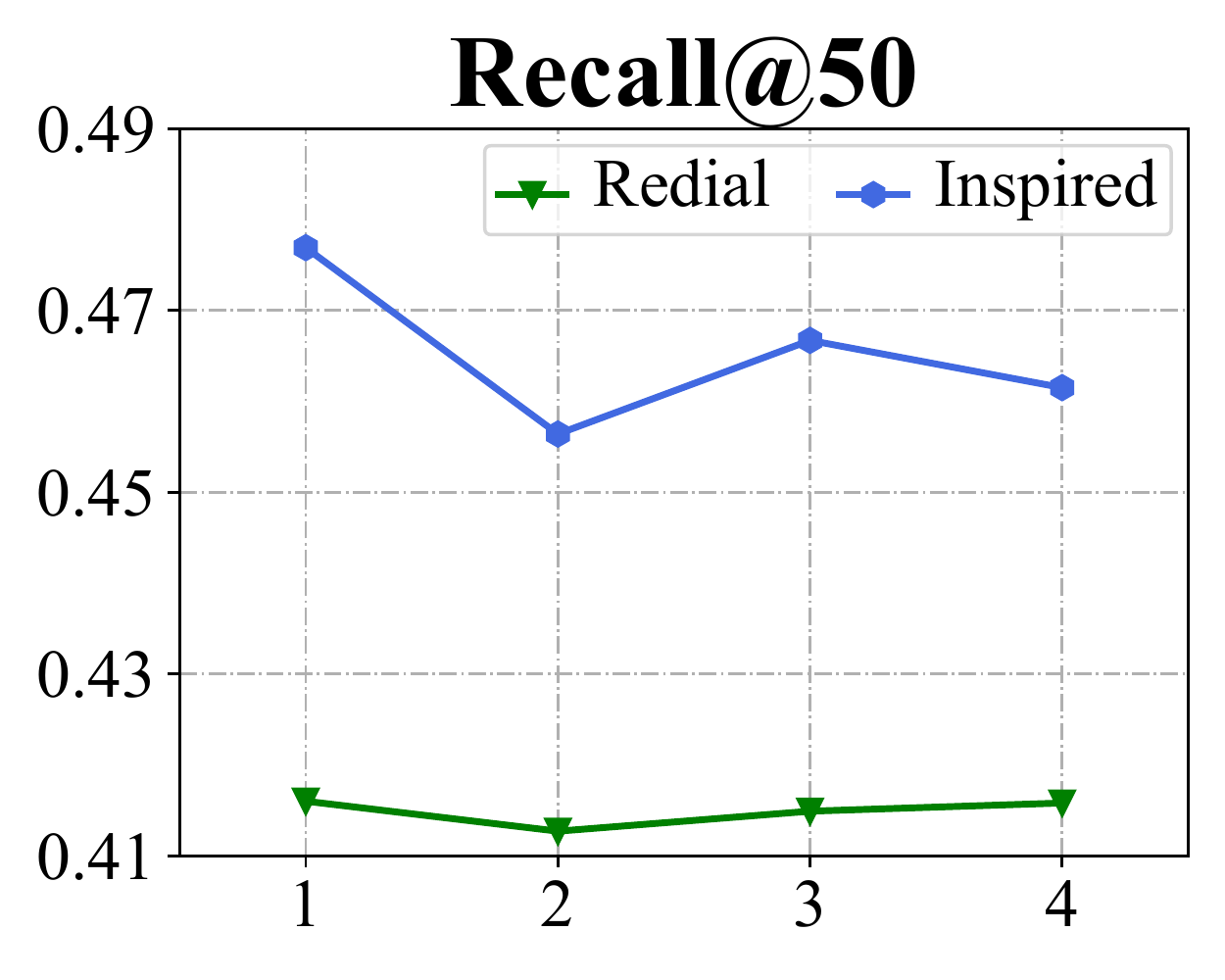}
     \includegraphics[width=0.49\linewidth]{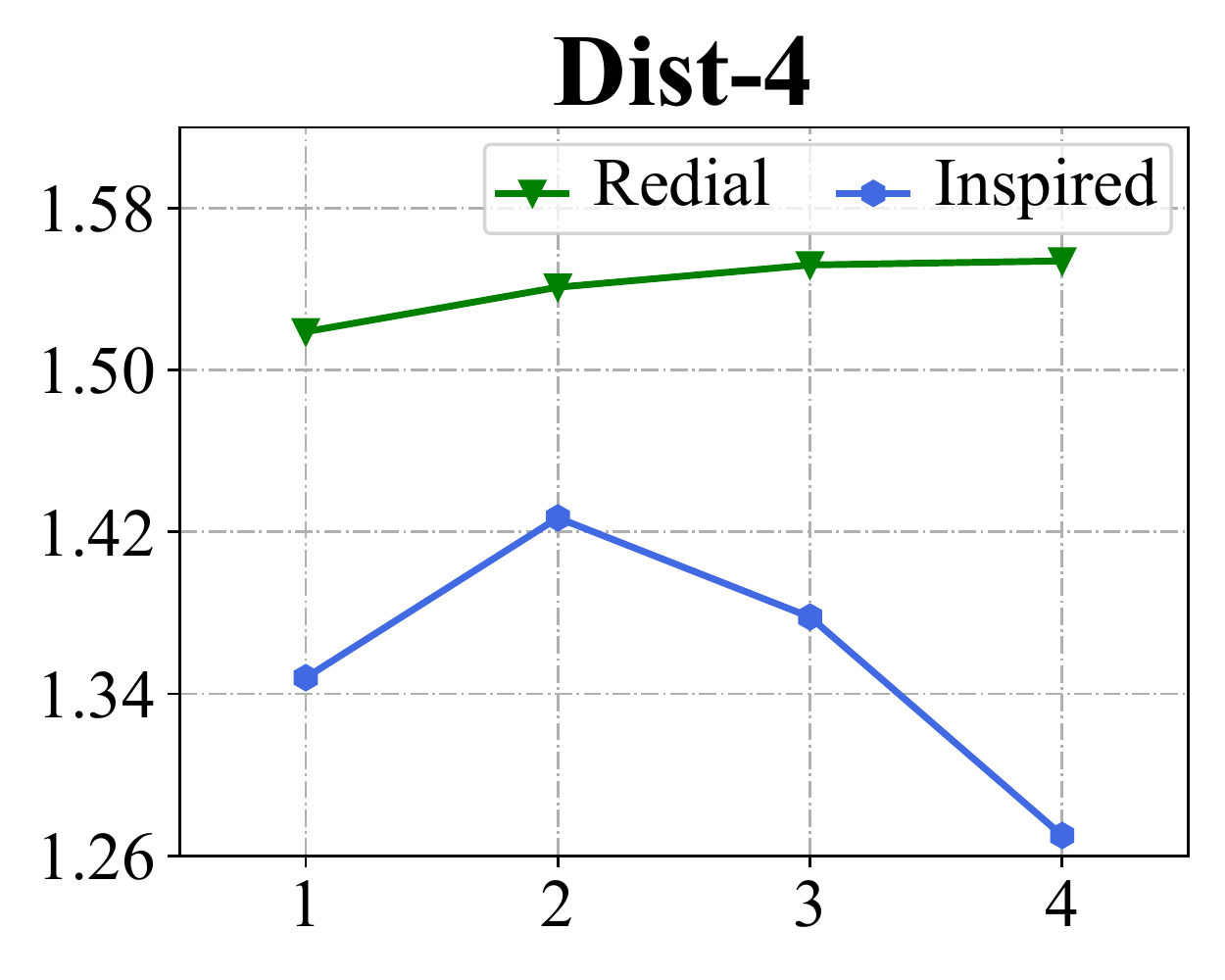}
     \caption{Performance comparison w.r.t. different ratios of augmented examples on \textsc{ReDial} and \textsc{Inspired} dataset. We implement our approach on KBRD.
     }
     \label{fig:para-num}
\end{figure}

\begin{figure}[t]
     \centering
     \includegraphics[width=0.49\linewidth]{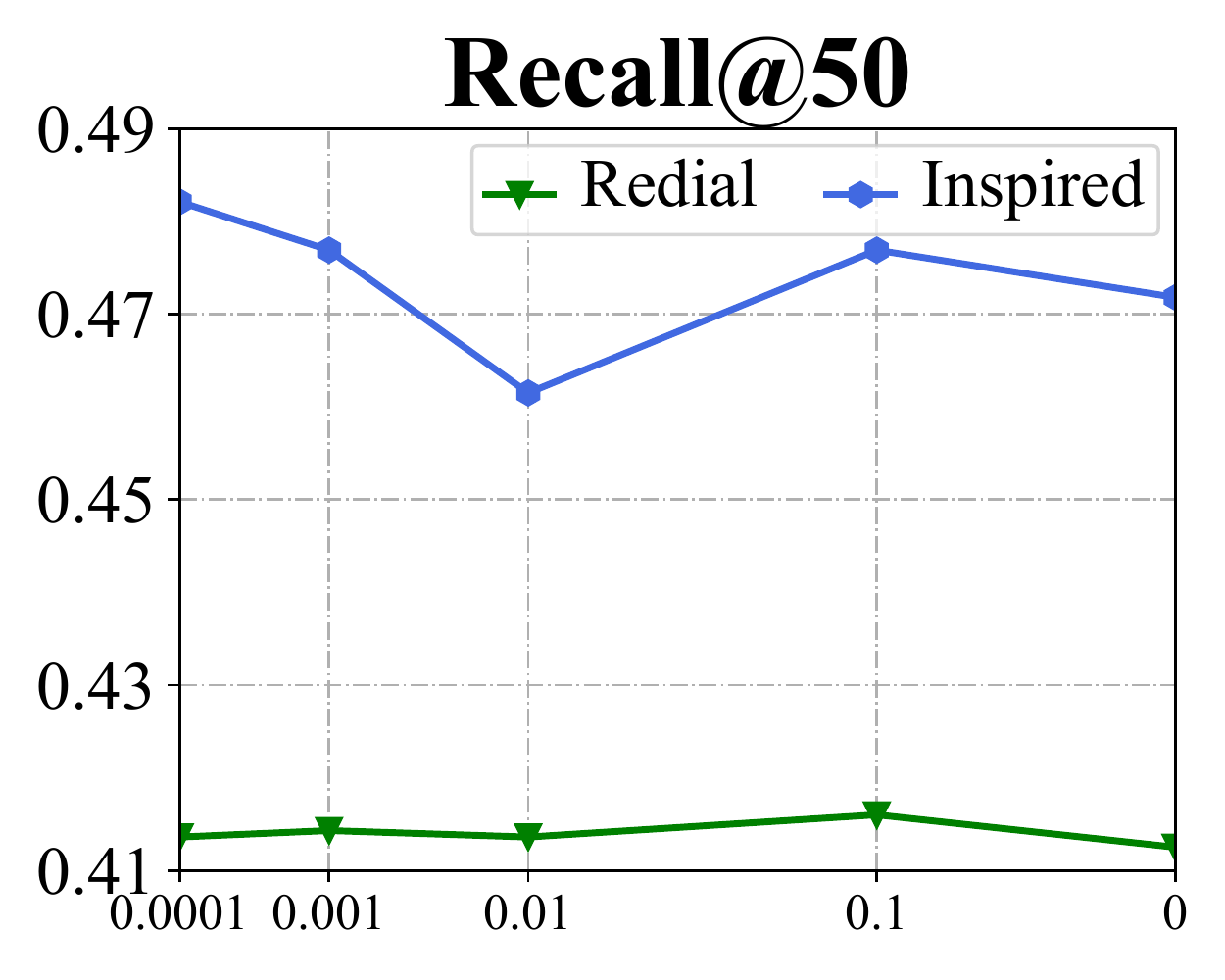}
     \includegraphics[width=0.49\linewidth]{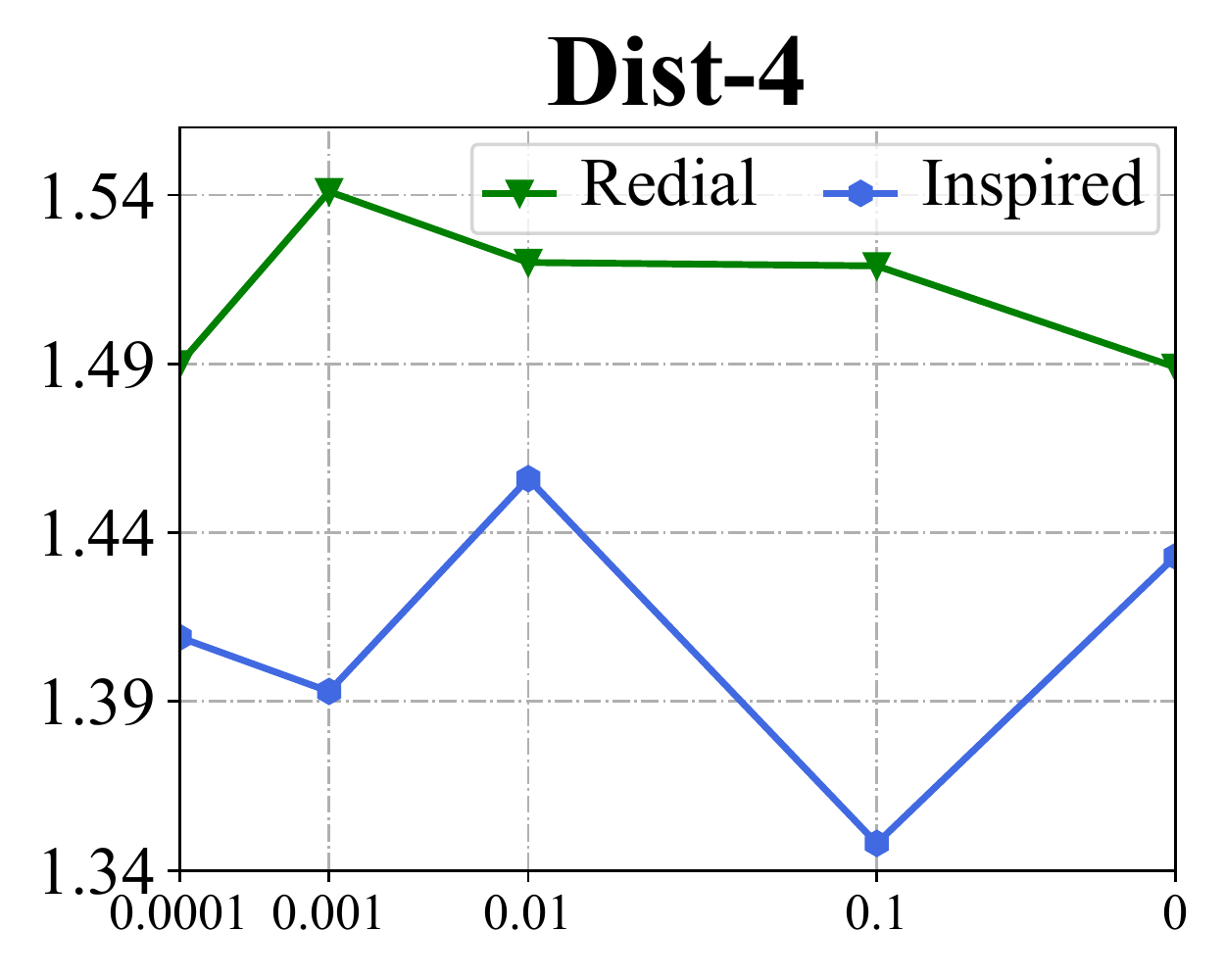}
     \caption{Performance comparison w.r.t. different weights of the L2-norm loss $\lambda$ on \textsc{ReDial} and \textsc{Inspired} dataset. We implement our approach on KBRD.
     }
     \label{fig:para-weight}
\end{figure}

In our framework, there are two major hyper-parameters to tune: the ratios of augmented examples for each instance and the weights of the L2-norm loss $\lambda$ during the adversarial training. 
Here, we investigate the effect of each hyper-parameter on our approach.
We conduct the analysis experiments on the recommendation and conversation tasks on the \textsc{ReDial} and \textsc{INSPIRED} datasets.
We implement our approach on KBRD and report the results for the two hyper-parameters in Figure~\ref{fig:para-num} and Figure~\ref{fig:para-weight}, respectively.

First, we can see that for the \textsc{ReDial} dataset, the performance is stable when tuning the two hyper-parameters.
It indicates that our approach is not too sensitive to the two hyper-parameters on this dataset.
Whereas, too large or too small weights of the L2-norm loss $\lambda$ would cause performance degradation, since too large $\lambda$ might punish the noise vectors too much, while too small one may bring too much noise.
Second, for the \textsc{INSPIRED} dataset, we can see the performance is not stable. 
A possible reason is that \textsc{INSPIRED} owns very few conversations in the training set, which may hurt the robustness of CRS models.
Besides, on the \textsc{INSPIRED} dataset, the best points of the two hyper-parameters in the recommendation and conversation tasks are different.
The reason may be that the two tasks focus on different goals, which may lead to conflict in their best hyper-parameter settings.

\section{Conclusion}
In this paper, we proposed a counterfactual data simulation approach, named \textbf{CFCRS}, for alleviating the issue of data scarcity in CRSs. 
We developed our approach under the framework of \textit{counterfactual data augmentation}, and employed counterfactual learning to enhance the quality of the augmented recommendation dialogue data. 
Specially, in our approach, we characterized the conversation flow and user preference via the entities mentioned in the conversation.
Our approach gradually augmented the user preference from a real dialogue without interfering with the entire conversation flow. 
Such an augmentation strategy was well learned by an adversarial training method with a curriculum schedule. 
As a key component, we designed a multi-stage recommendation dialogue simulator based on a conversation flow language model, which can generate \textit{reasonable, coherent} conversation flows for dialogue realization. 
Extensive experiments have shown that our approach can consistently boost the performance of several competitive CRSs, and outperform other data augmentation methods. 
   
Currently, our approach adopts a multi-stage stimulation method to generate recommendation dialogue data. 
For future work, we will investigate more unified and simplified approaches for high-quality data augmentation, such as the utilization of large language models~\cite{zhao2023survey, wang2023rethinking}.

\begin{acks}
    We are thankful to Jinhao Jiang for his supportive work.
    This work was partially supported by National Natural Science Foundation of China under Grant No. 62222215, Beijing Natural Science Foundation under Grant No. 4222027, Beijing Outstanding Young Scientist Program under Grant No. BJJWZYJH012019100020098, the Outstanding Innovative Talents Cultivation Funded Programs 2022 of Renmin University of China.
    Xin Zhao is the corresponding author.
\end{acks}

\bibliographystyle{ACM-Reference-Format}
\bibliography{ref}

\end{document}